\documentclass[10pt,letterpaper,twocolumn]{article}
\usepackage[latin9]{inputenc}
\usepackage{color}
\usepackage{array}
\usepackage{booktabs}
\usepackage{textcomp}
\usepackage{mathtools}
\usepackage{multirow}
\usepackage{amsmath}
\usepackage{amssymb}
\usepackage{stmaryrd}
\usepackage{graphicx}

\usepackage[unicode=true,
 bookmarks=false,
 breaklinks=true,pdfborder={0 0 1},backref=section,colorlinks=false]
 {hyperref}
\hypersetup{
	pagebackref=true,letterpaper=true,colorlinks}

\makeatletter

\pdfpageheight\paperheight
\pdfpagewidth\paperwidth

\newcommand{\noun}[1]{\textsc{#1}}
\providecommand{\tabularnewline}{\\}

\usepackage{cvpr}
\usepackage{times}
\usepackage{epsfig}
\usepackage{algorithm}
\usepackage{algpseudocode}\usepackage{amsfonts}

\usepackage{cuted}
\usepackage{capt-of}

\cvprfinalcopy 

\def\cvprPaperID{2829} 

\makeatother
\begin{document}



\title{Image Deformation Meta-Networks for One-Shot Learning}


\author{Zitian Chen$^1$\qquad Yanwei Fu$^1\thanks{Yanwei Fu is the corresponding author.}$\qquad Yu-Xiong Wang$^2$\qquad Lin Ma$^3$\qquad Wei Liu$^3$\qquad Martial Hebert$^2$\\$^1$Schools of Computer Science, and Data Science, Fudan University; Jilian Technology Group (Video++) \\ $^2$Robotics Institute, Carnegie Mellon University\qquad $^3$Tencent AI Lab\\  {\tt\small \{chenzt15,yanweifu\}@fudan.edu.cn\qquad yuxiongw@cs.cmu.edu\qquad forest.linma@gmail.com} \\ {\tt\small wl2223@columbia.edu\qquad hebert@cs.cmu.edu}}

\maketitle

\pagestyle{empty}  
\thispagestyle{empty} 

\begin{strip}
\centering
\includegraphics[trim={0cm 0cm -0cm 0cm},clip,height=.13\linewidth]{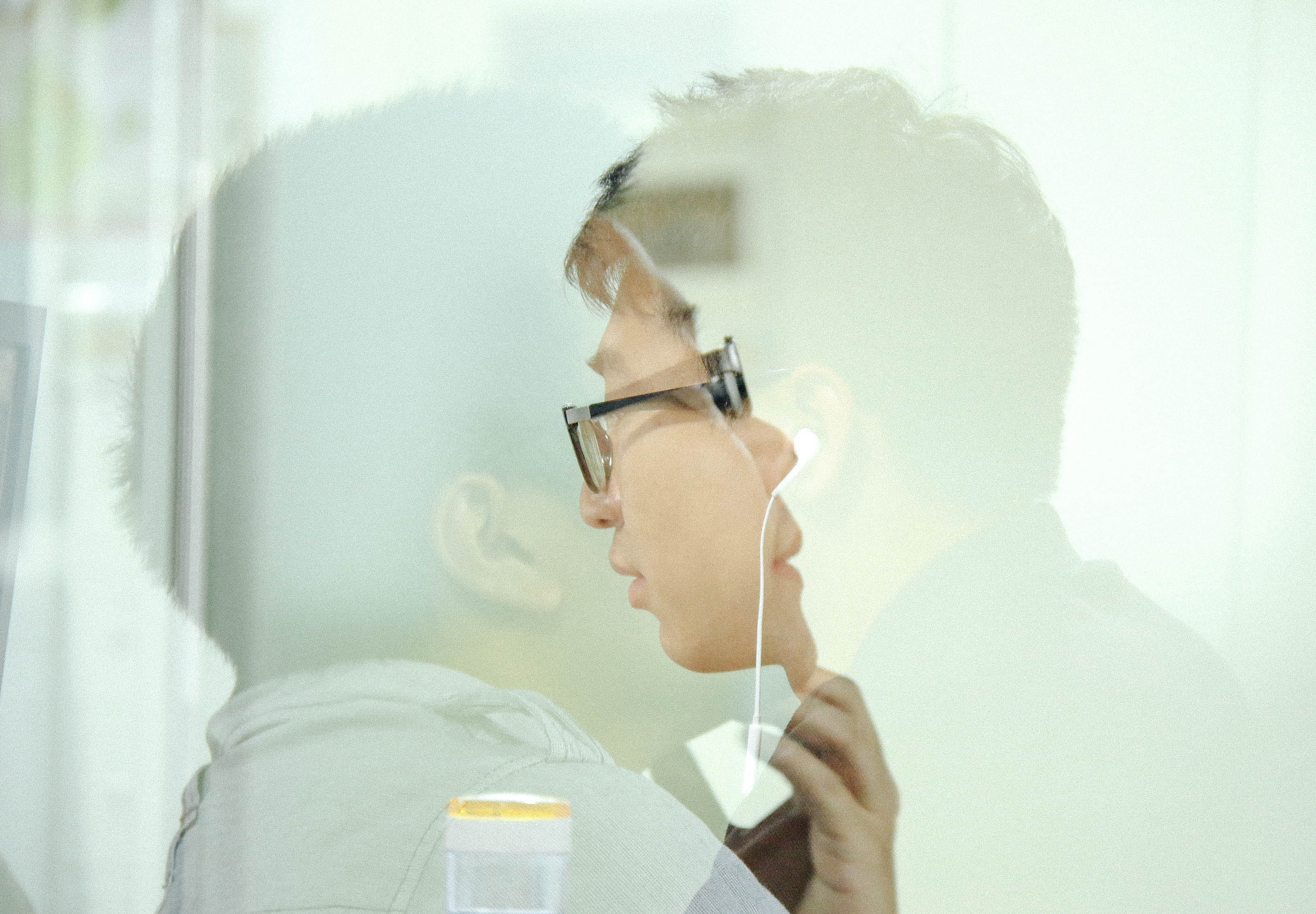}~
\includegraphics[trim={0cm 0cm 0cm 0cm},clip,height=.13\linewidth]{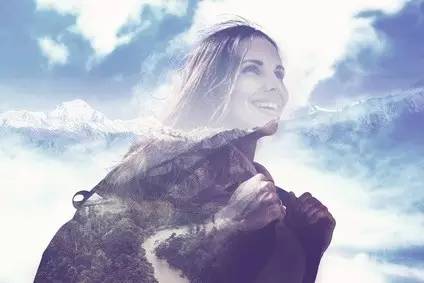}~
\includegraphics[trim={0cm 0cm 0cm 0cm},width=0.17\linewidth,clip,height=.13\linewidth]{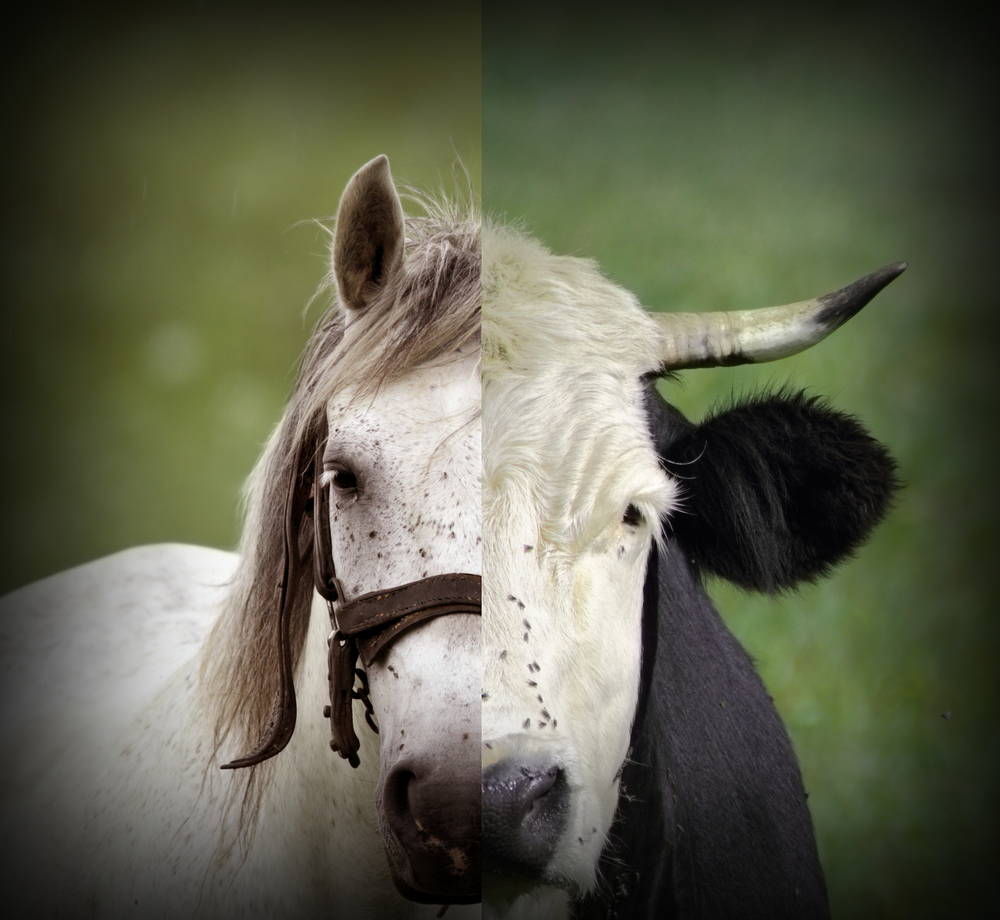}~
\includegraphics[trim={0cm 0cm 0cm 0cm},clip,height=.13\linewidth]{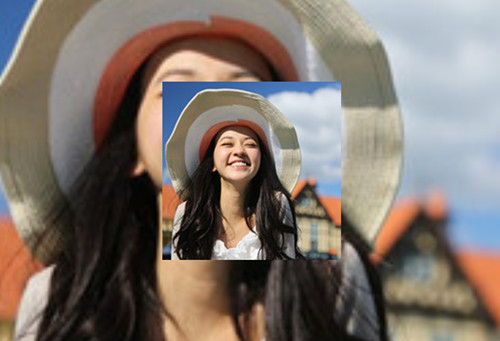}~
\includegraphics[trim={0cm 0cm 0cm 0cm},clip,height=.13\linewidth]{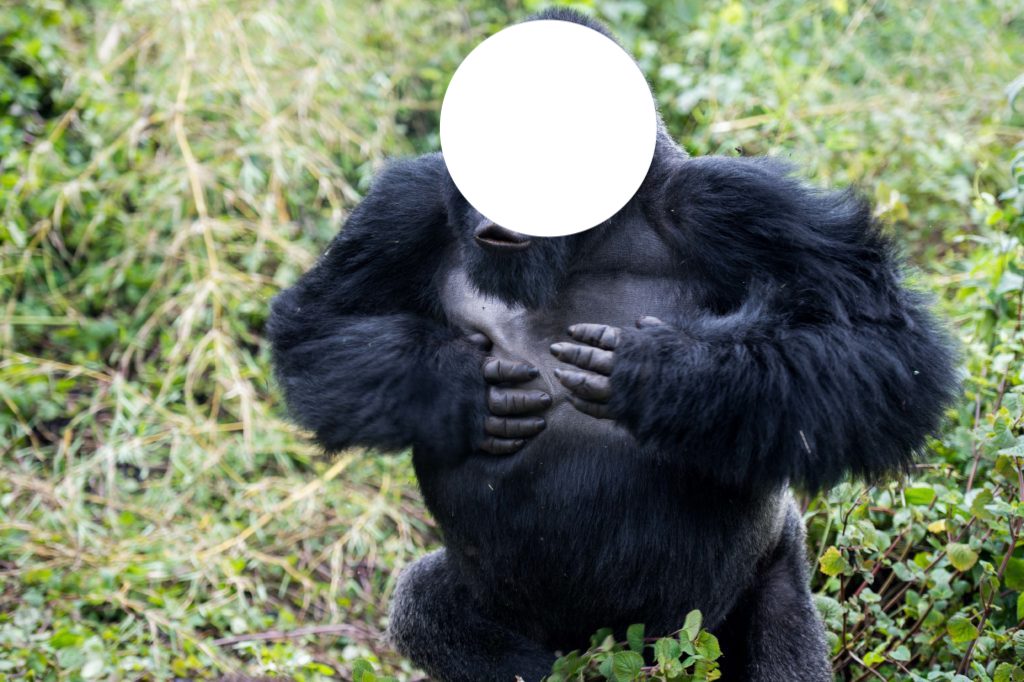}\\
\text{(a)}\qquad\qquad\qquad\quad\qquad
\text{(b)}\qquad\qquad\qquad\qquad\quad
\text{(c)}\qquad\qquad\qquad\quad\quad
\text{(d)}\qquad\qquad\qquad\quad\qquad
\text{(e)}
\vspace{0.05in}
\captionof{figure}{\label{fig:Illustration-of-human}Illustration of a variety of image deformations: ghosted (a, b), stitched (c), montaged (d), and partially occluded (e) images.}
\end{strip}

\begin{abstract}
Humans can robustly learn novel visual concepts even when images undergo various deformations and lose certain information. Mimicking the same behavior and synthesizing deformed instances of new concepts may help
visual recognition systems perform better one-shot learning, i.e.,
learning concepts from one or few examples. Our key insight is that,
while the deformed images may not be visually realistic, they still
maintain critical semantic information and contribute significantly
to formulating classifier decision boundaries. Inspired by the recent
progress of meta-learning, we combine a meta-learner with an image
deformation sub-network that produces additional training examples,
and optimize both models in an end-to-end manner. The deformation
sub-network learns to deform images by fusing a pair of images
--- a probe image that keeps the visual content and a gallery image
that diversifies the deformations. We demonstrate results on the widely
used one-shot learning benchmarks (miniImageNet and ImageNet
1K Challenge datasets), which significantly outperform state-of-the-art approaches. Code is available at \url{https://github.com/tankche1/IDeMe-Net}.

\end{abstract}

\vspace{-.7cm}
 
\section{Introduction}
Deep architectures have made significant progress in various visual
recognition tasks, such as image classification and object detection.
This success typically relies on supervised learning from large amounts
of labeled examples. In real-world scenarios, however, one may not
have enough resources to collect large training sets or need to deal
with rare visual concepts. It is also unlike the human visual system,
which can learn a novel concept with very little supervision.
One-shot or low/few-shot learning~\cite{feifei2006one_shot}, which aims to build a classifier for a new concept from
one or very few labeled examples, has thus attracted more and more
attention.

Recent efforts to address this problem have leveraged a \emph{learning-to-learn}
or \emph{meta-learning} paradigm ~\cite{Thrun1998,schmidhuber1987evolutionary,matchingnet_1shot,yuxiong2016NeurIPS,wang2016learningfrom,prototype_network,wang2017learning,Sachin2017,MAML,meta-sgd}.
Meta-learning algorithms train a learner, which is a parameterized function that maps labeled training
sets to classifiers. Meta-learners are trained by sampling a collection
of one-shot learning tasks and the corresponding datasets from a large
universe of labeled examples of known (base) categories, feeding the
sampled small training set to the learner to obtain a classifier,
and then computing the loss of the classifier on the sampled test
set. The goal is that the learner is able to tackle the recognition
of unseen (novel) categories from few training examples.

Despite their noticeable performance improvements, these generic meta-learning
algorithms typically treat images as black boxes and ignore the structure
of the visual world. By contrast, our biological vision system is
very robust and trustable in understanding images that undergo various \emph{deformations}~\cite{vermaak2005sensor}. For instance,
we can easily recognize the objects in Figure~\ref{fig:Illustration-of-human},
despite ghosting (Figure~\ref{fig:Illustration-of-human}(a, b)), stitching
(Figure~\ref{fig:Illustration-of-human}(c)), montaging (Figure~\ref{fig:Illustration-of-human}(d)),
and partially occluding (Figure~\ref{fig:Illustration-of-human}(e)) the images.
While these deformed images may not be visually realistic, \emph{our
key insight} is that they still maintain critical semantic information
and presumably serve as ``hard examples'' that contribute significantly
to formulating classifier decision boundaries. Hence, by leveraging
such modes of deformations shared across categories, the synthesized
deformed images could be used as additional training data to build
better classifiers.

A natural question then arises: how could we produce informative deformations?
We propose a simple parametrization
that linearly combines a pair of images to generate the deformed image.
We use a {\em probe} image to keep the visual content and overlay a {\em gallery}
image on a patch level to introduce appearance variations, which could
be attributed to semantic diversity, artifacts, or even random noise. Figure~\ref{fig:example} shows some examples of our deformed images. 
Importantly, inspired by~\cite{imaginaryData}, we \emph{learn to
deform }image\emph{s} that are useful for a classification objective by end-to-end
meta-optimization that includes image
deformations in the model.

Our \emph{Image Deformation Meta-Network} (IDeMe-Net) thus consists
of two components: a deformation sub-network and an embedding sub-network.
The deformation sub-network learns to generate the deformed images
by linearly fusing the patches of probe and gallery images. Specifically,
we treat the given small training set as the probe images and sample
additional images from the base categories to form the gallery images.
We evenly divide the probe and gallery images into nine patches, and
the deformation sub-network estimates the combination weight of each
patch. The synthesized images are used to augment the probe
images and train the embedding sub-network, which maps images to feature
representations and performs one-shot classification. The entire network
is trained in an end-to-end meta-learning manner on base categories.

\textbf{Our contributions} are three-fold. (1) We propose a novel
image deformation framework based on meta-learning to address one-shot
learning, which leverages the rich structure of shared modes of deformations
in the visual world. (2) Our deformation network learns to synthesize diverse deformed images, which effectively exploits the
complementarity and interaction between the probe and gallery image
patches. (3) By using the deformation network, we effectively augment
and diversify the one-shot training images, leading to a significant
performance boost on one-shot learning tasks. Remarkably, our approach
achieves state-of-the-art performance on both the challenging
ImageNet1K and \emph{mini}ImageNet datasets.


\section{Related Work}


\noindent \textbf{Meta-Learning.} Typically, meta-learning~\cite{Thrun1998,Thrun96learningto,schmidhuber1987evolutionary,matchingnet_1shot,yuxiong2016NeurIPS,wang2016learningfrom,prototype_network,wang2017learning,Sachin2017,MAML,meta-sgd,DEML+Meta-SGD,MetaNetwork} aims at training a parametrized mapping from a few training instances
to model parameters in simulated one-shot learning scenarios.
Other meta-learning strategies in one-shot learning include graph CNNs~\cite{2017arXiv171104043G} and memory networks~\cite{deep_1shot_recent,memorymatching}. Attention is also introduced in meta-learning, in ways of analyzing the
relation between visual and semantic representations~\cite{multiAttention} and
learning the combination of temporal convolutions and soft attention~\cite{SNAIL}. Different from prior work,
we focus on exploiting the complementarity
and interaction between visual patches through the meta-learning mechanism.

\noindent \textbf{Metric Learning.} This is another important line of work in one-shot learning. The goal is to learn
a metric space which can be optimized for one-shot learning. Recent
work includes Siamese networks~\cite{siamese_1shot}, matching
networks~\cite{matchingnet_1shot}, prototypical networks~\cite{prototype_network}, relation
networks~\cite{relation_net}, and dynamic few-shot learning without forgetting~\cite{dym}.

\noindent \textbf{Data Augmentation.} The key limitation of one-shot
learning is the lack of sufficient training images. As a common practice, data augmentation has been widely used to help train supervised classifiers~\cite{KrizhevskySH12,returnDevil2014BMVC,visualizing_network}.
The standard techniques include adding Gaussian noise, flipping, rotating, rescaling, transforming, and randomly cropping training images. However, the generated images in this way are particularly subject to visual similarity with the original images. In addition to adding noise or jittering, previous work seeks to augment training
images by using semi-supervised techniques~\cite{wang2016learningfrom,ren18fewshotssl,ladderNet} and utilizing relation between visual and semantic representations~\cite{semanticAugmentation} ,
or directly synthesizing
new instances in the feature domain~\cite{2017ICCVaug,imaginaryData,Delta-encoder,cogan} to transfer knowledge of data distribution from base classes to
novel classes. By contrast, we also use samples from base classes
to help synthesize deformed images but directly aim at maximizing
the one-shot recognition accuracy.

The most relevant to our approach is the work of~\cite{imaginaryData,mixup}. Wang \etal~\cite{imaginaryData} introduce a GAN-like generator
to hallucinate novel instances in the feature domain by adding noise, whereas we focus
on learning to deform two real images {\em in the image domain without introducing noise}. Zhang \etal~\cite{mixup} randomly sample image pairs and linearly combine them to generate additional training images. In this {\em mixup} augmentation, the combination is performed with weights randomly sampled from a prior distribution and is thus constrained to be convex. The label of the generated image is similarly the linear combination of the labels (as one-hot label vectors) of the image pairs. However, they ignore structural dependencies between images as well as image patches. By contrast, we  learn classifiers to select images that are similar to the probe images from the {\em unsupervised} gallery image set. Our combination weights are learned through a deformation sub-network {\em on the image patch level} and the combination {\em is not necessarily convex}. In addition, our generated image preserves the label of its probe image. Comparing
with these methods, our approach learns to dynamically fuse patches
of two real images in an end-to-end manner. The produced images
maintain the important patches of original images while being visually different from them, thus facilitating training one-shot classifiers.


\section{One-Shot Learning Setup}


\label{sec:oneShotSetting}

Following recent work~\cite{matchingnet_1shot,Sachin2017,MAML,prototype_network,imaginaryData},
we establish one-shot learning in a \emph{meta-learning} framework:
we have a base category set $\mathcal{C}_{base}$ and a novel category
set $\mathcal{C}_{novel}$, in which $\mathcal{C}_{base}\cap\mathcal{C}_{novel}=\emptyset$;
correspondingly, we have a base dataset $D_{base}=\left\{ \left(\mathbf{I}_{i},y_{i}\right), y_{i}\in\mathcal{C}_{base}\right\} $ and a novel dataset $D_{novel}=\left\{ \left(\mathbf{I}_{i},y_{i}\right), y_{i}\in\mathcal{C}_{novel}\right\}$. We aim to learn a classification algorithm
on $D_{base}$ that can generalize to unseen classes $C_{novel}$
with one or few training examples per class.

To mimic the one-shot learning scenario, meta-learning algorithms
learn from a collection of \emph{$N$-way-$m$-shot} classification
tasks/datasets sampled from $D_{base}$ and are evaluated in a similar
way on $D_{novel}$. Each of these sampled datasets is termed as \emph{an
episode}, and we thus have different meta-sets for meta-training and
meta-testing. Specifically, we randomly sample $N$ classes $L\sim C_{k}$
for a meta-training (\emph{i.e.}, $k=base$) or meta-testing episode
(\emph{i.e.}, $k=novel$). We then randomly sample $m$ and $q$ labeled
images per class in $L$ to construct the support set $S$ and query
set $Q$, respectively, \emph{i.e.}, $|S|=N\times m$ and $|Q|=N\times q$.
During meta-training, we sample $S$ and $Q$ to train our model.
During meta-testing, we evaluate by averaging the classification accuracy
on query sets $Q$ of many meta-testing episodes.

We view the support set as supervised probe images and different from
the previous work, we introduce an additional gallery image set $G$
that serves as an \emph{unsupervised} image pool to help generate
deformed images. To construct $G$, we randomly sample some images
per base class from \emph{the base dataset}, \emph{i.e.}, $G\sim D_{base}$.
The same $G$ is used in both the meta-training and meta-testing episodes.
Note that since it is purely sampled from $D_{base}$, the newly introduced
$G$ does not break the standard one-shot setup as in~\cite{prototype_network,MAML,Sachin2017}.
We do not introduce any additional images from the novel categories
$\mathcal{C}_{novel}$.

\section{Image Deformation Meta-Networks}

\begin{figure*}

\begin{centering}
\begin{tabular}{c}
\includegraphics[scale=0.19]{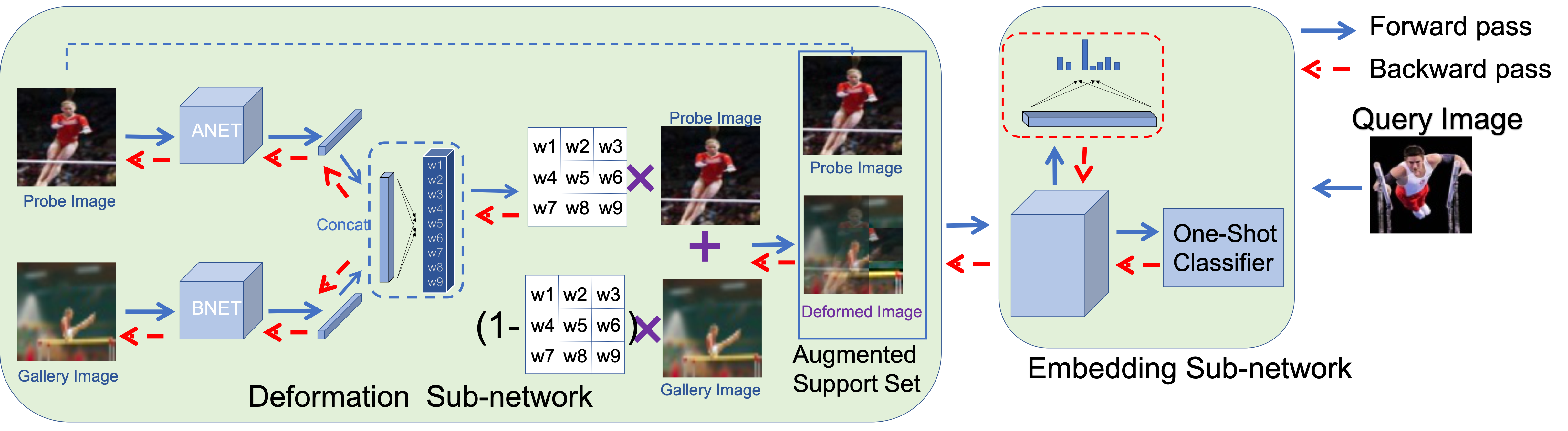}\tabularnewline
\end{tabular}
\par\end{centering}
\caption{\label{fig:FuseNet}The overall architecture of our image deformation
meta-network (IDeMe-Net). }

 
\end{figure*}

We now explain our image deformation meta-network (IDeMe-Net) for
one-shot learning. Figure \ref{fig:FuseNet} shows the architecture
of IDeMe-Net $f_{\theta}(\cdot)$ parametrized by $\theta$. IDeMe-Net
is composed of two modules --- a deformation sub-network and an embedding
sub-network. The deformation sub-network adaptively fuses the probe
and gallery images to synthesize the deformed images. The embedding
sub-network maps the images to feature representations and then constructs
the one-shot classifier. The entire meta-network is trained in an
end-to-end manner.

\subsection{Deformation Sub-network}

This sub-network $f_{\theta_{def}}\left(\cdot\right)$ learns to explore
the interaction and complementarity between the probe images $\mathbf{I}_{probe}$
($\left( \mathbf{I}_{probe},y_{probe}\right) \in S$) and the gallery
images $\mathbf{I}_{gallery}\in G$, and fuses them to generate the
synthesized deformed images $\mathbf{I}_{syn}$, \emph{i.e.}, $\mathbf{I}_{syn}=f_{\theta_{def}}\left(\mathbf{I}_{probe},\mathbf{I}_{gallery}\right)$.
Our goal is to synthesize meaningful deformed images such that
$y_{syn}=y_{probe}$. This is achieved by using two strategies: (1)
$y_{syn}=y_{probe}$ is explicitly enforced as a constraint during
the end-to-end optimization; (2) we propose an approach
to sample $\mathbf{I}_{gallery}$ that are visually or semantically
similar to the images of $y_{probe}$. Specifically, for each class $y_{probe}$, we directly use
the feature extractor and one-shot classifier learned in the embedding
sub-network to select the top $\epsilon\%$ images from $G$ which
have the highest class probability of $y_{probe}$. From this initial pool of images, we randomly sample $\mathbf{I}_{gallery}$ for each probe image $\left( \mathbf{I}_{probe},y_{probe}\right)$. Note that during meta-training, both $\mathbf{I}_{probe}$ and $\mathbf{I}_{gallery}$ are randomly sampled from base classes, so they might belong to the same class. We find that further constraining them to belong to different base classes has little impact on the performance. During meta-testing, $\mathbf{I}_{probe}$ and $\mathbf{I}_{gallery}$ belong to different classes, with $\mathbf{I}_{probe}$ sampled from novel classes and $\mathbf{I}_{gallery}$ still from base classes.



Two branches, ANET and BNET, are used to parse $\mathbf{I}_{probe}$
and $\mathbf{I}_{gallery}$, respectively. Each of them is a residual
network \cite{he2015deep} without fully-connected layers. The outputs
of ANET and BNET are then concatenated to be fed into a fully-connected
layer, which produces a 9-D weight vector $w$. As shown in Figure~\ref{fig:FuseNet}, we evenly divide the images into $3\times3$ patches. The deformed
image is thus simply generated as a linearly weighted combination of $\mathbf{I}_{probe}$ and $\mathbf{I}_{gallery}$ on the patch level. That is, for the $q$th patch, we have

\noindent 
\begin{equation}
\mathbf{I}_{syn,q}=w_q\mathbf{I}_{probe,q}+\left(1-w_q\right)\mathbf{I}_{gallery,q}.\label{eq:synthesize-1}
\end{equation}

We assign
the class label $y_{probe}$ to the synthesized deformed image $\mathbf{I}_{syn}$.\textcolor{red}{{}
}For any probe image $\mathbf{I}_{probe}^{i}$, we sample $n_{aug}$
gallery images from the corresponding pool and produce $n_{aug}$ synthesized
deformed images. We thus
obtain an augmented support set 
\noindent 
\begin{equation}
\hspace{-0.1in}
\tilde{S}=\left\{ \left( \mathbf{I}^{i}_{probe},y^{i}_{probe}\right),\left\{ \left(\mathbf{I}_{syn}^{i,j},y_{probe}^{i,j}\right)\right\}_{j=1}^{n_{aug}}\right\}_{i=1}^{N\times m}.\label{eq:aug_set}
\end{equation}


\subsection{Embedding Sub-network }
The embedding sub-network $f_{\theta_{emb}}(\cdot)$
consists of a deep convolutional network for feature extraction and
a non-parametric one-shot classifier.
Given an input image $\mathbf{I}$, we use a residual network \cite{he2015deep}
to produce its feature representation $f_{\theta_{emb}}\left(\mathbf{I}\right)$.
To facilitate the training process, we introduce an additional softmax classifier, \ie, a fully-connected
layer on top of the embedding sub-network with a cross-entropy loss
(CELoss), that outputs $|\mathcal{C}_{base}|$ scores. 

\subsection{One-Shot Classifier}

Due to its superior performance, we use the non-parametric prototype
classifier \cite{prototype_network} as the one-shot classifier. During
each episode, given the sampled $S$, $Q$, and $G$, the deformation
sub-network produces the augmented support set $\tilde{S}$. Following
\cite{prototype_network}, we calculate the prototype vector $p_{\theta}^{c}$
for each class $c$ in $\tilde{S}$ as

\begin{equation}
p_{\theta}^{c}=\frac{1}{Z}\sum_{(\mathbf{I}_{i},y_{i})\in\tilde{S}}f_{\theta_{emb}}\left(\mathbf{I}_{i}\right)\cdot\left\llbracket y_{i}=c\right\rrbracket ,\label{eq:p_thetat}
\end{equation}

\noindent where $Z=\Sigma_{(\mathbf{I}_{i},y_{i})\in\tilde{S}}\left\llbracket y_{i}=c\right\rrbracket $
is the normalization factor. $\left\llbracket \cdot\right\rrbracket $
is the Iverson's bracket notation: $\left\llbracket x\right\rrbracket =1$
if $x$ is true, and $0$ otherwise. Given any query image $\mathbf{I}_{i}\in Q$,
its probability of belonging to class $c$ is computed as

\noindent 
\begin{equation}
\hspace{-0.1in}
P_{\theta}\left(y_{i}=c\text{\ensuremath{\mid}}\mathbf{I}_{i}\right)=\frac{\mathrm{exp}\left(-\left\Vert \mathit{f}_{\theta_{emb}}\left(\mathbf{I}_{i}\right)-\mathit{p}_{\theta}^{c}\right\Vert^{2} \right)}{\sum_{j=1}^{N}\mathrm{exp}\left(-\left\Vert f_{\theta_{emb}}\left(\mathbf{I}_{i}\right)-p_{\theta}^{j}\right\Vert^{2} \right)},\label{eq:prototypical_classifier}
\end{equation}

\noindent where $\parallel\cdot\parallel$ indicates the Euclidean
distance. The one-shot classifier $P$ thus predicts the class label of
$\mathbf{I}_{i}$ as the highest probability over $N$ classes.

\section{Training Strategy of IDeMe-Net}

\begin{algorithm}    	 
\caption{Meta-training procedure of our IDeMe-Net $f_{\theta}$. $G$ is the fixed gallery constructed from $\mathcal{C}_{base}$.}\label{alg}    \begin{algorithmic}[1]   	 
\Procedure{Meta-Train\_Episode}{} \\The procedure of one meta-training episode	
	 \State $L\gets$ randomly sample $N$ classes from $\mathcal{C}_{base}$       	
	 \State $S\gets$ randomly sample instances belonging to $L$ \State{//sample the support set}     		
\State $Q\gets$ randomly sample instances belonging to $L$  \State{//sample the query set}   
     		 \State train the prototype classifier $P$ from $f_{\theta_{emb}}(S)$    		 
\State $\tilde{S}\gets S$ \Comment{initialize the augment support set}       		
\For{$c$ in $L$}\Comment{enumerate the chosen classes}      	 			 
\State $pool\gets$use $P$ to select $\epsilon\% $ images in $G$ that have the highest class probability of $c$          		 
	\For{$(\mathbf{I}_{probe},c)$ in $S_c$}    
		\For {$j=1$ to $n_{aug}$}    	 			 	
\State $\mathbf{I}_{gallery}\gets $ randomly sample instances from $pool$              			 	
\State $\mathbf{I}_{syn}\gets f_{\theta_{def}}(\mathbf{I}_{probe},\mathbf{I}_{gallery})$               			 	
\State $\tilde{S} \gets \tilde{S}\cup (\mathbf{I}_{syn},c)$        
		\EndFor{}  		 
	\EndFor{}       		 
\EndFor{}              	
	\State train the prototype classifier $\tilde{P}$ from $f_{\theta_{emb}}(\tilde{S})$       	 
	\State use $\tilde{P}$ to classify $f_{\theta_{emb}}(Q)$ and obtain the prototype loss       	
	\State use the softmax classifier to classify $f_{\theta_{emb}}(\tilde{S})$ and obtain the CELoss       	
	\State update $\theta_{emb}$ with the CELoss       		
\State update $\theta_{def}$ with the prototype loss                    
	 	\EndProcedure     	
\end{algorithmic} 

\end{algorithm}

\subsection{Training Loss}
Training the entire IDeMe-Net includes two subtasks: (1) training
the deformation sub-network which maximally improves the one-shot
classification accuracy; (2) building the robust embedding sub-network
which effectively deals with various synthesized deformed images.
Note that our one-shot classifier has no parameters, which does not
need to be trained. We use the prototype loss and the cross-entropy
loss to train these two sub-networks, respectively.

\noindent \textbf{Update the deformation sub-network.} We optimize
the following prototype loss function to endow the deformation
sub-network with the desired one-shot classification ability:
\begin{equation}
\hspace{-0.3cm}\mathrm{min}_{\theta}\mathbb{E}_{G,L\sim D_{base}}\mathbb{E}_{S,Q\sim L}\left[\sum_{\left(\mathbf{I}_{i},y_{i}\right)\in Q}-\mathrm{log}P_{\theta}\left(y_{i}\mid\mathbf{I}_{i}\right)\right],\label{eq:loss_function}
\end{equation}
\noindent where $P_{\theta}\left(y_{i}\mid\mathbf{I}_{i}\right)$
is the one-shot classifier in Eq.~(\ref{eq:prototypical_classifier}).
Using the prototype loss encourages the deformation sub-network to generate diverse instances to augment the support set.

\noindent \textbf{Update the embedding sub-network. }We use the cross-entropy
loss to train the embedding sub-network to directly classify the augmented
support set $\tilde{S}$. Note that with the augmented support set
$\tilde{S}$, we have relatively more training instances to train this
sub-network and the cross-entropy loss is the standard loss function
in training a supervised classification network. Empirically, we find
that using the cross-entropy loss speeds up the convergence and improves
the recognition performance than using the prototype loss only.

\subsection{Training Strategy}

We summarize the entire training procedure of our IDeMe-Net
on the base dataset $D_{base}$ in Algorithm~\ref{alg}. We have access to the {\em same, predefined} gallery $G$ from $D_{base}$ for both meta-training and meta-testing. During meta-training,
we sample the $N$-way-$m$-shot training
episode to produce $S$ and $Q$ from $D_{base}$. The embedding sub-network learns
an initial one-shot classifier on $S$ using Eq.~(\ref{eq:prototypical_classifier}). Given
a probe image $\mathbf{I}_{probe}$, we then sample the gallery images
$\mathbf{I}_{gallery}\sim G$ and train the deformation sub-network
to generate the augmented support set $\tilde{S}$ using Eq.~(\ref{eq:synthesize-1}).
$\tilde{S}$ is further used to update the embedding sub-network and
learn a better one-shot classifier. We then conduct the final one-shot
classification on the query set $Q$ and back-propagate the prediction
error to update the entire network. During meta-testing, we sample
the $N$-way-$m$-shot testing episode to produce $S$ and $Q$ from
the novel dataset $D_{novel}$.

\section{Experiments}

\begin{table*}
\begin{centering}
\begin{tabular}{llccccc}
\hline 
 & {\small{}{}{}{}{}{}Method } & {\small{}{}{}{}{}{}$m=1$  } & {\small{}{}{}{}{}{}$2$  } & {\small{}{}{}{}{}{}$5$  } & {\small{}{}{}{}{}{}$10$  } & {\small{}{}{}{}{}{}$20$}\tabularnewline
\hline 
\hline 
\multirow{4}{*}{{\small{}Baselines }} & {\small{}{}{}{}{}{}Softmax  } & {\small{}{}{}{}{}{}-- / 16.3  } & {\small{}{}{}{}{}{}-- / 35.9  } & {\small{}{}{}{}{}{}-- / 57.4  } & {\small{}{}{}{}{}{}-- / 67.3  } & {\small{}{}{}{}{}{}-- / 72.1}\tabularnewline
 & {\small{}{}{}{}{}{}LR  } & {\small{}18.3{}/42.8  } & {\small{}{}{}{}{}{}{}26.0{}/54.7  } & {\small{}{}{}{}{}{}{}35.8{}/66.1  } & {\small{}{}{}{}{}{}{}41.1{}/71.3  } & {\small{}{}{}{}{}{}{}44.9{}/74.8 }\tabularnewline
 & {\small{}{}{}{}{}{}SVM  } & {\small{}{}{}{}{}{}15.9/36.6  } & {\small{}{}{}{}{}{}22.7/48.4  } & {\small{}{}{}{}{}{}31.5/61.2  } & {\small{}{}{}{}{}{}37.9/69.2  } & {\small{}{}{}{}{}{}43.9/74.6}\tabularnewline
 & {\small{}{}{}{}{}{}Prototype Classifier  } & {\small{}{}{}{}{}{}17.1/39.2  } & {\small{}{}{}{}{}{}24.3/51.1  } & {\small{}{}{}{}{}{}33.8/63.9  } & {\small{}{}{}{}{}{}38.4/69.9  } & {\small{}{}{}{}{}{}44.1/74.7 }\tabularnewline
\hline 
\hline 
\multirow{7}{*}{{\small{}Competitors }} 
& {\small{}{}{}{}{}{}Matching Network~\cite{matchingnet_1shot}
 } & {\small{}{}{}{}{}{}-- / 43.0  } & {\small{}{}{}{}{}{}-- / 54.1  } & {\small{}{}{}{}{}-- / 64.4  } & {\small{}{}{}{}{}{}-- / 68.5  } & {\small{}{}{}{}{}{}-- / 72.8 }\tabularnewline
 & {\small{}{}{}Prototypical Network~\cite{prototype_network} } & {\small{}{}{}{}{}{}16.9/41.7  } & {\small{}{}{}{}{}{}24.0/53.6  } & {\small{}{}{}{}{}{}33.5/63.7  } & {\small{}{}{}{}{}{}37.7/68.2  } & {\small{}{}{}{}{}{}42.7/72.3}\tabularnewline
 & {\small{}{}{}{}{}{}Generation SGM~\cite{2017ICCVaug} } & {\small{}{}{}{}{}{}-- / 34.3  } & {\small{}{}{}{}{}{}-- / 48.9  } & {\small{}{}{}{}{}{}-- / 64.1  } & {\small{}{}{}{}{}{}-- / 70.5  } & {\small{}{}{}{}{}{}-- / 74.6}\tabularnewline
 & {\small{}{}{}PMN~\cite{imaginaryData} } & {\small{}{}{}-- / 43.3  } & {\small{}{}{}-- / 55.7  } & {\small{}{}{}-- / 68.4  } & {\small{}{}{}-- / 74.0  } & {\small{}{}{}-- / 77.0}\tabularnewline
 & {\small{}{}{}PMN w/ H~\cite{imaginaryData} } & {\small{}{}{}-- / 45.8  } & {\small{}{}{}-- / 57.8  } & {\small{}{}{}-- / 69.0  } & {\small{}{}{}-- / 74.3  } & {\small{}{}{}-- / 77.4}\tabularnewline
 & {\small{}{}{}Cos \& Att.~\cite{dym} } & {\small{}{}{}-- / 46.0  } & {\small{}{}{}-- / 57.5  } & {\small{}{}{}-- / 69.1  } & {\small{}{}{}-- / 74.8}{\small{}{}  } & {\small{}{}{}-- / 78.1}\tabularnewline
 & {\small{}{}CP-AAN~\cite{cogan} } & {\small{}{}-- / {}48.4 } & {\small{}{}-- / {}59.3 } & {\small{}{}-- / {}70.2 } & {\small{}{}-- / {}\textbf{\small{}{}{}{}76.5} } &  {\small{}{}{}-- / }\textbf{\small{}{}{}{}79.3}\tabularnewline
\hline 
\multirow{4}{*}{{\small{}Augmentation }} & {\small{}{}{}{}{}{}Flipping  } & {\small{}{}{}{}{}{}17.4/39.6  } & {\small{}{}{}{}{}{}24.7/51.2  } & {\small{}{}{}{}{}{}33.7/64.1  } & {\small{}{}{}{}{}{}38.7/70.2  } & {\small{}{}{}{}{}{}44.2/74.5 }\tabularnewline
 & {\small{}{}{}{}{}{}Gaussian Noise  } & {\small{}{}{}{}{}{}16.8/39.0  } & {\small{}{}{}{}{}{}24.0/51.2  } & {\small{}{}{}{}{}{}33.9/63.7  } & {\small{}{}{}{}{}{}38.0/69.7  } & {\small{}{}{}{}{}{}43.8/74.5 }\tabularnewline
 & {\small{}{}{}Gaussian Noise (feature level)  } & {\small{}{}{}{}{}{}16.7/39.1  } & {\small{}{}{}{}{}{}24.2/51.4  } & {\small{}{}{}{}{}{}33.4/63.3  } & {\small{}{}{}{}{}{}38.2/69.5  } & {\small{}{}{}{}{}{}44.0/74.2 }\tabularnewline
 & {\small{}{}{}{}{}{}Mixup~\cite{mixup}} & {\small{}{}{}{}{}{}15.8/38.7  } & {\small{}{}{}{}{}{}24.6/51.4  } & {\small{}{}{}{}{}{}32.0/61.1  } & {\small{}{}{}{}{}{}38.5/69.2  } & {\small{}{}{}{}{}{}42.1/72.9 }\tabularnewline
\hline 
\hline 
{\small{}Ours  } & {\small{}{}{}{}{}{}IDeMe-Net  } & \textbf{\small{}{}{}{}{}{}{}23.1/51.0}{\small{}{}{}{}  } & \textbf{\small{}{}30.1/60.9}{\small{}{}  } & \textbf{\small{}{}{}{}{}{}39.3/70.4}{\small{}{}  } & \textbf{\small{}{}42.7{}/}{\small{}{}73.4  } & \textbf{\small{}{}{}{}{}{}45.0{}/}{\small{}{}{}75.1}\tabularnewline
\hline 
\end{tabular}
\vspace{0.1in}
\caption{\label{tab:Imagenet1k-resnet10}\textbf{Top-1 / Top-5 accuracy (\%)
on novel classes of the ImageNet 1K Challenge dataset.} We use \textbf{ResNet-10}
as the embedding sub-network. $m$ indicates the number of training
examples per class.  Our IDeMe-Net consistently achieves the best performance.}
\par\end{centering}
 
\end{table*}

Our IDeMe-Net is evaluated on two standard benchmarks: \emph{mini}ImageNet~\cite{matchingnet_1shot}
and ImageNet 1K Challenge~\cite{2017ICCVaug} datasets. \emph{mini}ImageNet
is a widely used benchmark in one-shot learning, which includes 600
images per class and has 100 classes in total. Following the data
split in \cite{Sachin2017}, we use 64, 16, 20 classes as the base,
validation, and novel category set, respectively. The hyper-parameters
are cross-validated on the validation set. Consistent with \cite{matchingnet_1shot,Sachin2017}, we evaluate our model in 5-way-5-shot and
5-way-1-shot settings.

For the large-scale ImageNet 1K dataset, we divide the original
1K categories into 389 base ($D_{base}$) and 611 novel ($D_{novel}$)
classes following the data split in \cite{2017ICCVaug}. The base classes are further divided into two disjoint
subsets: base validation set $D_{base}^{cv}$ (193 classes) and
evaluation set $D_{base}^{fin}$ (196 classes) and the novel classes
are divided into two subsets as well: novel validation set $D_{novel}^{cv}$
(300 classes) and evaluation set $D_{novel}^{fin}$ (311 classes).
We use the base/novel validation set $D^{cv}$ for cross-validating
hyper-parameters and use the base/novel evaluation set $D^{fin}$
to conduct the final experiments. The same experimental setup is used
in~\cite{2017ICCVaug} and the reported results are averaged over
5 trials. Here we focus on synthesizing novel instances and we thus
evaluate the performance primarily on novel classes, \ie,
311-way-$m$-shot settings, which is also consistent with most of the
contemporary work \cite{matchingnet_1shot,prototype_network,Sachin2017}.

\noindent 
\begin{table}
\begin{centering}
{\small{}}%
 \resizebox{\linewidth}{!}{
\begin{tabular}{c}
$\hspace{-0.45cm}$%
\begin{tabular}{lcccc}
\toprule 
{\small{}Method } & {\small{}{}{}{}{}{}$m=1$ } & {\small{}{}{}{}{}{}$2$ } & {\small{}{}{}{}{}{}$5$ } & {\small{}{}{}{}{}{}$10$ }\tabularnewline
\midrule 
{\small{}{}{}{}{}Softmax } & {\small{}{}{}{}{}-- / 28.2 } & {\small{}{}{}{}{}-- / 51.0 } & {\small{}{}{}{}{}-- / 71.0 } & {\small{}{}{}{}{}-- / 78.4 }\tabularnewline
{\small{}{}{}{}{}SVM } & {\small{}{}{}{}{}20.1/41.6 } & {\small{}{}{}{}{}29.4/57.7 } & {\small{}{}{}{}{}42.6/72.8 } & {\small{}{}{}{}{}49.9/79.1 }\tabularnewline
{\small{}{}{}{}{}LR } & {\small{}{}{}{}{}{}22.9{}/47.9 } & {\small{}{}{}{}{}{}32.3{}/61.3 } & {\small{}{}{}{}{}{}44.3{}/73.6 } & {\small{}{}{}{}{}{}50.9{}/78.8 }\tabularnewline
Proto-Clsf & {\small{}{}{}{}{}20.8/43.1 } & {\small{}{}{}{}{}29.9/58.1 } & {\small{}{}{}{}{}42.4/72.3 } & {\small{}{}{}{}{}49.5/79.0 }\tabularnewline
\midrule
\midrule 
{\small{}{}G-SGM~\cite{2017ICCVaug}} & {\small{}{}{}{}{}-- / 47.3 } & {\small{}{}{}{}{}-- / 60.9 } & {\small{}{}{}{}{}-- / 73.7 } & {\small{}{}{}{}{}-- / 79.5 }\tabularnewline
{\small{}{}PMN~\cite{imaginaryData} } & {\small{}{}-- / 53.3 } & {\small{}{}-- / 65.2 } & {\small{}{}-- / 75.9 } & {\small{}{}-- / 80.1}\tabularnewline
{\small{}{}PMN w/ H~\cite{imaginaryData} } & {\small{}{}-- / {}54.7 } & {\small{}{}-- / {}66.8 } & {\small{}{}-- /  {}\textbf{77.4} } & {\small{}{}-- / {}\textbf{81.4}}\tabularnewline
{\small{}{}IDeMe-Net (Ours)} & \textbf{\small{}{}{}{}{}{}30.3{}/{}60.1}{\small{}{}{}{} } & \textbf{\small{}{}{}{}{}{}39.7{}/{}69.6}{\small{}{}{}{} } & \textbf{\small{}{}{}{}{}{}47.5{}/{}77.4}{\small{}{}{}{} } & \small{}{}{}{}{}{}\textbf{51.3}{}/{}80.2{\small{}{}{}{} }\tabularnewline
\bottomrule
\end{tabular}\tabularnewline
\end{tabular}}
\vspace{0.1in}
\caption{\label{tab:Imagenet1k-resnet50}\textbf{Top-1 / Top-5 accuracy (\%)
on novel classes of the Imagenet 1K Challenge dataset.} We use \textbf{ResNet-50}
as the embedding sub-network. $m$ indicates the number of training
examples per class. Proto-Clsf and G-SGM denote the prototype
classifier and generation SGM~\cite{2017ICCVaug}, respectively.}
\par\end{centering}
\end{table}


\subsection{Results on ImageNet 1K Challenge\label{subsec:Results-on-ImageNet1k}}

\noindent \textbf{Setup.} We use ResNet-10 architectures for ANET
and BNET (\ie, the deformation sub-network). For a fair comparison with \cite{2017ICCVaug,imaginaryData}, we evaluate
the performance of using ResNet-10 (Table \ref{tab:Imagenet1k-resnet10})
and ResNet-50 (Table \ref{tab:Imagenet1k-resnet50}) for the embedding
sub-network. Stochastic gradient descent (SGD) is used to train IDeMe-Net
in an end-to-end manner. It gets converged over 100 epochs. The initial
learning rates of ANET, BNET, and the embedding sub-network are set as
$3\times10^{-3}$, $3\times10^{-3}$, and $10^{-1}$, respectively,
and decreased by $1/10$ every 30 epochs. The batch size is set as
32. We randomly sample 10 images per base category to construct the
gallery $G$ and we set $\epsilon$ as 2. Note that $G$ is \emph{fixed}
during the entire experiments. ANET, BNET, and the embedding sub-network
are trained from scratch on $D_{base}$. Our model is evaluated on
$D_{novel}$. $n_{aug}$ is cross-validated as 8, which balances between
the computational cost and the augmented training data scale. In practice, we perform stage-wise training to overcome potential negative influence caused by misleading training images synthesized by the initial deformation sub-network. Specifically, to make the training more robust, we first fix the deformation sub-network and train the embedding sub-network with real and deformed images. Here the deformed images are synthesized by linearly combining two images on a patch level with a randomly sampled weight vector $w$. Note that these two images are sampled from the same category. Then we fix the embedding sub-network and learn the deformation sub-network to reduce the discrepancy between synthesized and real images. Finally, we train the embedding and deformation sub-networks jointly (\ie, the entire IDeMe-Net) to allow them to cooperate with each other.

\noindent \textbf{Baselines and Competitors.} We compare against several
baselines and competitors as follows. (1) We directly train a ResNet-10
feature extractor on $D_{base}$ and use it to compute  features
on $D_{novel}$. We then train standard supervised classifiers on
$D_{novel}$, including neural network, support vector machine
(SVM), logistic regression (LR), and prototype classifiers. The neural
network classifier consists of a fully-connected layer and a softmax
 layer. (2) We compare with state-of-the-art approaches
to one-shot learning, such as matching networks~\cite{matchingnet_1shot},
generation SGM \cite{2017ICCVaug}, prototypical networks~\cite{prototype_network},
Cosine Classifier \& Att. Weight Gen (Cos \& Att.)~\cite{dym}, CP-ANN~\cite{cogan},
PMN, and PMN w/ H~\cite{imaginaryData}. (3) The data augmentation
methods are also compared --- flipping: the input image is
flipped from left to right; Gaussian noise: cross-validated
Gaussian noise $\mathcal{N}\left(0,10\right)$ is added to each pixel
of the input image; Gaussian noise (feature level): cross-validated
Gaussian noise $\mathcal{N}\left(0,0.3\right)$ is added to each dimension
of the ResNet feature for each image; Mixup: using mixup~\cite{mixup} to combine probe and gallery images. For fair comparisons, all these
augmentation methods use the prototype classifier as the one-shot
classifier.
\begin{table*}
\begin{centering}
\begin{tabular}{llccccc}
\hline 
 & {\small{}{}{}{}{}{}Method  } & {\small{}{}{}{}{}{}$m=1$  } & {\small{}{}{}{}{}{}$2$  } & {\small{}{}{}{}{}{}$5$  } & {\small{}{}{}{}{}{}$10$  } & {\small{}{}{}{}{}{}$20$ }\tabularnewline
\hline 
\hline 
\multirow{2}{*}{{\small{}Baselines }} & {\small{}{}{}{}{}{}LR  } & {\small{}{}{}{}{}{}18.3/42.8  } & {\small{}{}{}{}{}{}26.0/54.7  } & {\small{}{}{}{}{}{}35.8/66.1  } & {\small{}{}{}{}{}{}41.1/71.3  } & {\small{}{}{}{}{}{}44.9/74.8 }\tabularnewline
 & {\small{}{}{}{}{}{}Prototype Classifier  } & {\small{}{}{}{}{}{}17.1/39.2  } & {\small{}{}{}{}{}{}24.3/51.1  } & {\small{}{}{}{}{}{}33.8/63.9  } & {\small{}{}{}{}{}{}38.4/69.9  } & {\small{}{}{}{}{}{}44.1/74.7 }\tabularnewline
\hline 
\multirow{7}{*}{{\small{}Variants }} & {\small{}{}{}{}{}{}IDeMe-Net - CELoss  } & {\small{}{}{}{}{}{}21.3/50.0  } & {\small{}{}{}{}{}{}28.0/58.3  } & {\small{}{}{}{}{}{}37.7/69.4  } & {\small{}{}{}{}{}{}41.3/71.6  } & {\small{}{}{}{}{}{}44.3/74.3 }\tabularnewline
 & {\small{}{}{}{}{}IDeMe-Net - Proto Loss  } & {\small{}{}{}{}{}{}15.3/36.7  } & {\small{}{}{}{}{}{}21.4/50.4  } & {\small{}{}{}{}{}{}31.7/62.0  } & {\small{}{}{}{}{}{}37.9/69.0  } & {\small{}{}{}{}{}{}43.7/73.7}\tabularnewline
 & {\small{}{}{}{}{}{}IDeMe-Net - Predict
 } & {\small{}{}{}{}{}{}17.0/39.3  } & {\small{}{}{}{}{}{}24.0/50.7  } & {\small{}{}{}{}{}{}33.6/63.5  } & {\small{}{}{}{}{}{}38.0/69.2  } & {\small{}{}{}{}{}{}43.7/73.8 }\tabularnewline
 & {\small{}{}{}{}{}{}IDeMe-Net - Aug. Testing  } & {\small{}{}{}{}{}{}17.0/39.1  } & {\small{}{}{}{}{}{}24.30/51.3  } & {\small{}{}{}{}{}{}33.5/63.8  } & {\small{}{}{}{}{}{}38.0/69.1  } & {\small{}{}{}{}{}{}43.8/74.5 }\tabularnewline
 & {\small{}{}{}{}{}{}IDeMe-Net - Def. Network  } & {\small{}{}{}{}{}{}15.9/38.0  } & {\small{}{}{}{}{}{}24.1/50.1  } & {\small{}{}{}{}{}{}32.6/63.3  } & {\small{}{}{}{}{}{}38.2/68.9  } & {\small{}{}{}{}{}{}42.4/73.1 }\tabularnewline
 & {\small{}{}{}{}{}IDeMe-Net - Gallery  } & {\small{}{}{}{}{}17.5/39.4  } & {\small{}{}{}{}{}24.2/51.4  } & {\small{}{}{}{}{}33.5/63.7  } & {\small{}{}{}{}{}38.7/70.3  } & {\small{}{}{}{}{}44.4/74.5 }\tabularnewline
 & {\small{}{}{}{}{}IDeMe-Net - Deform  } & {\small{}{}{}{}{}15.7/37.8  } & {\small{}{}{}{}{}22.7/49.8  } & {\small{}{}{}{}{}31.9/62.6  } & {\small{}{}{}{}{}38.0/68.7  } & {\small{}{}{}{}{}43.5/73.8}\tabularnewline
\hline 
\multirow{5}{*}{{\small{}Patch Size }} 
& {\small{}{}{}{}{}{}IDeMe-Net ($1\times1$)  } & {\small{}{}{}{}{}{}16.2/39.3  } & {\small{}{}{}{}{}{}24.4/52.1  } & {\small{}{}{}{}{}{}32.9/63.0  } & {\small{}{}{}{}{}{}38.8/69.5  } & {\small{}{}{}{}{}{}42.7/73.2 }\tabularnewline
 & {\small{}{}{}{}{}{}IDeMe-Net ($5\times5$)  } & \textbf{\small{}{}{}{}{}{}{}24.1}{\small{}{}{}{}{}{}/{}51.7
 } & \textbf{\small{}{}{}{}{}{}}{\small{}{}{}{}{}{}30.3{}{}/{}61.2
 } & \textbf{\small{}{}{}{}{}{}{}39.6}{\small{}{}{}{}{}{}/}\textbf{\small{}{}{}{}{}{}{}70.4}{\small{}{}{}{}{}
 } & {\small{}{}{}{}{}{}42.4/{}73.2{}  } & {\small{}{}{}{}{}{}44.3/74.6}\tabularnewline
 & {\small{}{}{}{}{}{}IDeMe-Net ($7\times7$)  } & {\small{}{}{}{}{}{}{}23.8}\textbf{\small{}{}{}{}{}{}/{}52.1}{\small{}{}{}{}{}
 } & {\small{}{}{}{}{}{}30.2/}\textbf{\small{}{}{}{}{}{}{}61.3}{\small{}{}{}{}{}
 } & {\small{}{}{}{}{}{}39.1/70.2  } & {\small{}{}{}{}{}{}}\textbf{\small{}{}{}{}{}{}42.7}{\small{}{}{}{}{}/73.1
 } & \textbf{\small{}{}{}{}{}}{\small{}{}44.5{}/{}74.7}\tabularnewline
 & {\small{}{}{}{}{}{}IDeMe-Net (pixel level)  } & {\small{}{}{}{}{}{}17.3/39.0  } & {\small{}{}{}{}{}{}23.8/51.2  } & {\small{}{}{}{}{}{}34.1/63.7  } & {\small{}{}{}{}{}{}38.5/70.2  } & {\small{}{}{}{}{}{}43.9/74.5 }\tabularnewline
\hline 
{\small{}{}Ours  } & {\small{}{}{}{}{}{}IDeMe-Net  } & {\small{}{}{}{}{}23.1/51.0  } & \textbf{\small{}{}30.4{}/}{\small{}{}60.9  } & {\small{}{}{}{}{}{}39.3{}/{}70.4{}  } & \textbf{\small{}{}42.7/73.4}{\small{}{}  } & \textbf{\small{}{}{}{}{}{}45.0/75.1}\tabularnewline
\hline 
\end{tabular}
\par\end{centering}
\vspace{0.1in}
\caption{\textbf{\label{tab:resnet-10-imagenet1kablation}Top-1 / Top-5 accuracy (\%)
of the ablation study on novel classes of the ImageNet 1K Challenge dataset.}
We use ResNet-10 as the embedding sub-network. $m$ indicates the number of training
examples per class. Our full model achieves the best performance.}

 
\end{table*}

\noindent \textbf{Results.} Tables~\ref{tab:Imagenet1k-resnet10}
and~\ref{tab:Imagenet1k-resnet50} summarize the results of using
ResNet-10 and ResNet-50 as the embedding sub-network, respectively.
For example, using ResNet-10, the top-5 accuracy of IDeMe-Net
in Table~\ref{tab:Imagenet1k-resnet10} is superior to the prototypical
network by 7\% when $m=1,2,5$, showing the sample efficiency of IDeMe-Net
for one-shot learning. With more data (\eg, $m=10,20$), 
while the plain prototype classifier baseline performs worse than other baselines (\eg, PMN), 
our deformed images coupled with the prototype classifier still have significant effect (\eg, $3.5$ point boost when $m=10$).
The top-1 accuracy 
demonstrates the similar trend.
Using ResNet-50 as the embedding sub-network, the performance of all
the approaches improves and our IDeMe-Net consistently achieves the
best performance, as shown in Table~\ref{tab:Imagenet1k-resnet50}. Figure~\ref{fig:Ablation}(a) further highlights that our IDeMe-Net
consistently outperforms all the baselines by large
margins.


\subsection{Ablation Study on ImageNet 1K Challenge}

We conduct extensive ablation studies to evaluate the contribution of
each component in our model.

\begin{figure}
\begin{centering}
\begin{tabular}{cc}
$\hspace{-0.4cm}$\includegraphics[scale=0.3]{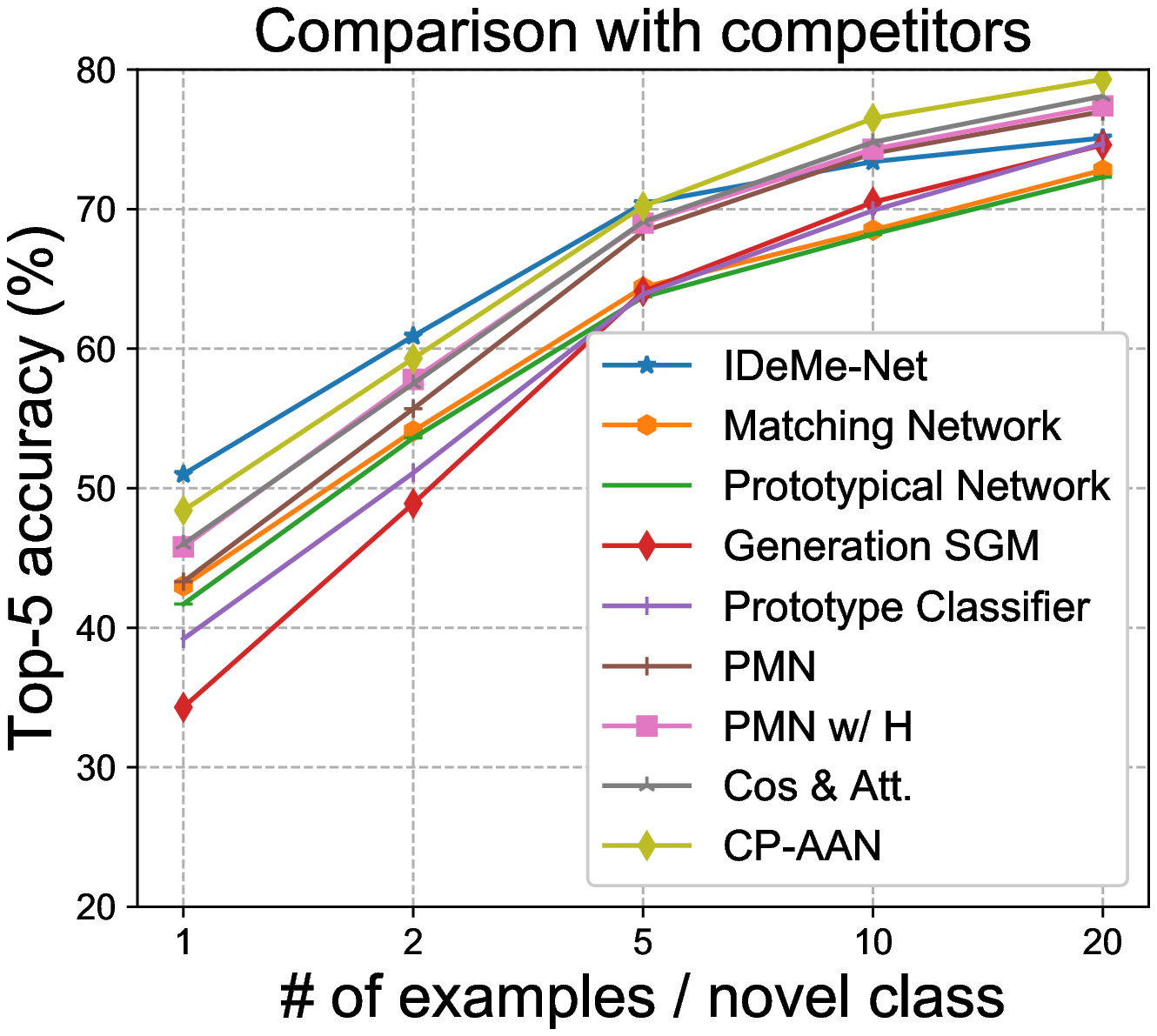}
$\hspace{-0.4cm}$  & $\hspace{-0.4cm}$\includegraphics[scale=0.3]{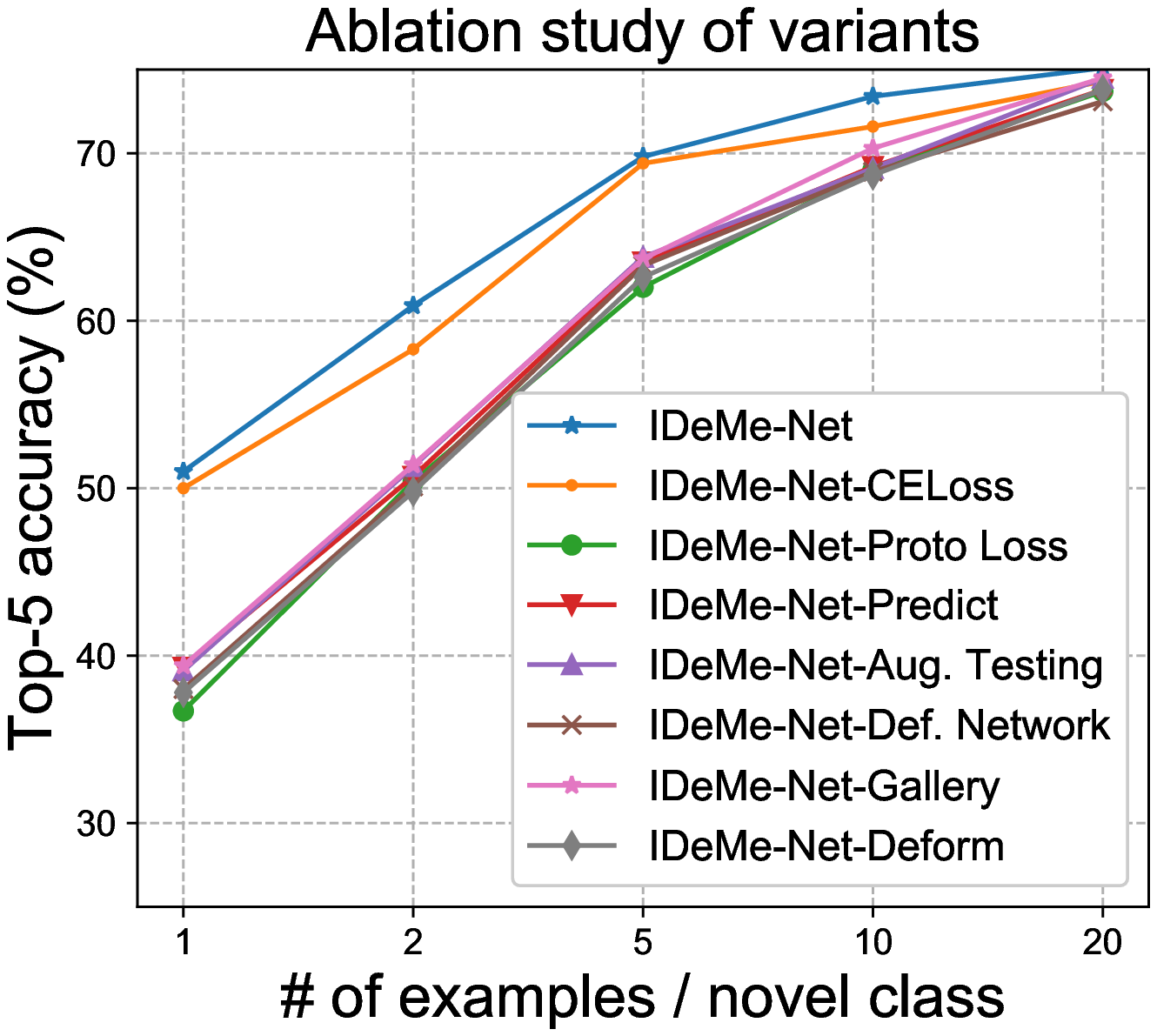} $\hspace{-0.4cm}$ \tabularnewline
{\small{}{}(a) }  & {\small{}{}(b) }\tabularnewline
$\hspace{-0.4cm}$\includegraphics[scale=0.3]{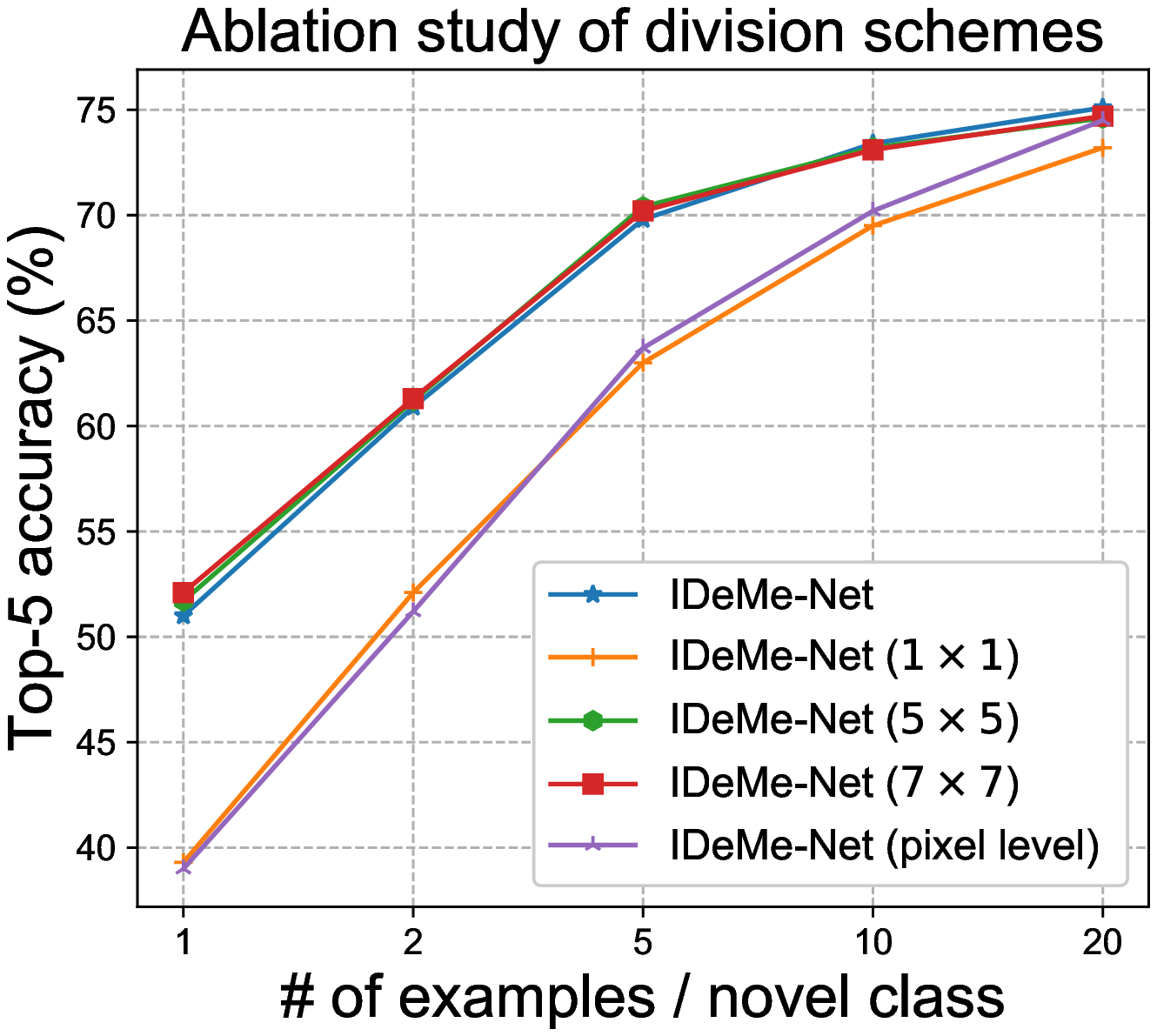}$\hspace{-0.4cm}$  & $\hspace{-0.4cm}$\includegraphics[scale=0.3]{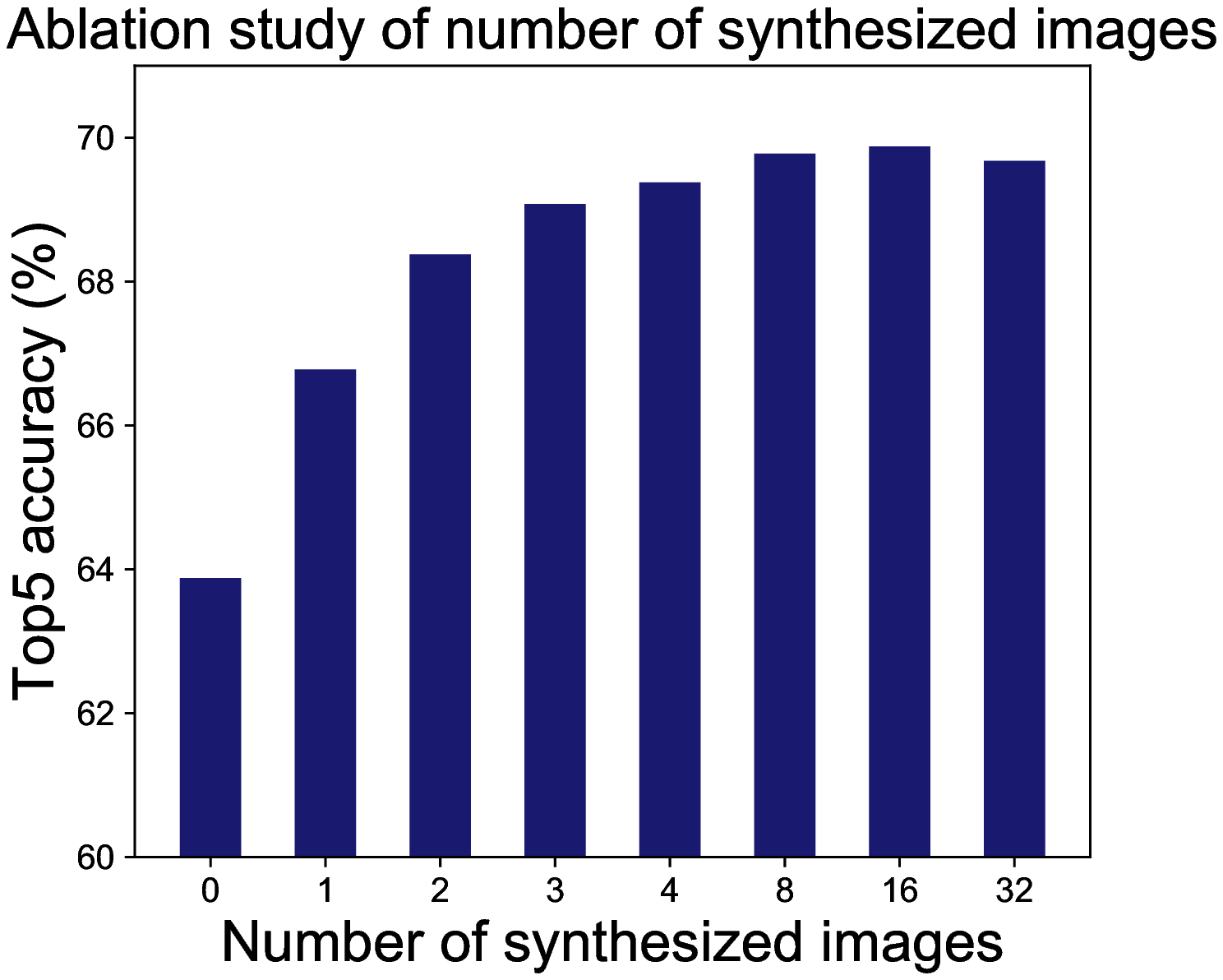}\tabularnewline
{\small{}{}(c) }  & {\small{}{}(d)}\tabularnewline
\end{tabular}
\par\end{centering}
\vspace{0.1in}
\caption{\textbf{\label{fig:Ablation}Ablation study on ImageNet 1K Challenge
dataset:} (a) highlights the comparison with several competitors;
(b) shows the impact of different components on our IDeMe-Net; (c)
analyzes the impact of different division schemes; (d) shows how the
performance changes with respect to the number of synthesized deformed
images. \textbf{Best viewed in color with zoom. }}

\end{figure}

\noindent \textbf{Variants of IDeMe-Net.} We consider seven different
variants of our IDeMe-Net, as shown in Figure~\ref{fig:Ablation}(b)
and Table~\ref{tab:resnet-10-imagenet1kablation}. (1) `IDeMe-Net - CELoss': the IDeMe-Net is trained using only the prototype loss
\emph{without} the cross-entropy loss (CELoss). (2) `IDeMe-Net - Proto
Loss': the IdeMe-Net is trained using only the cross-entropy
loss \emph{without} the prototype loss. (3) `IDeMe-Net - Predict':
the gallery images are randomly chosen in IDeMe-Net \emph{without}
predicting their class probability. (4) `IDeMe-Net - Aug. Testing':
the deformed images are not produced in the meta-testing phase. (5) `IDeMe-Net
- Def. Network': the combination weights in Eq.~(\ref{eq:synthesize-1})
are randomly generated instead of using the learned deformation sub-network.
(6) `IDeMe-Net - Gallery': the gallery images are directly sampled from
the support set instead of constructing an additional Gallery. (7)
`IDeMe-Net - Deform': we simply use the gallery images to serve as the
deformed images. As shown in Figure~\ref{fig:Ablation}(b), our full
IDeMe-Net model outperforms all these variants, showing that
each component is essential and complementary to each other.

We note that (1) \textbf{Using CELoss and prototype loss to update
the embedding and deformation sub-networks, respectively,
achieves the best result.} As shown in Figure~\ref{fig:Ablation}(b),
the accuracy of `IDeMe-Net - CELoss' is marginally lower than IDeMe-Net
but still higher than the prototype classifier baseline, while
`IDeMe-Net - Proto Loss' underperforms the baseline. (2) \textbf{Our
strategy for selecting the gallery images is the key to diversify
the deformed images.} Randomly choosing the gallery
images (`IDeMe-Net - Predict') or sampling
the gallery images from the support set (`IDeMe-Net - Gallery') obtains
no performance improvement. One potential explanation is that they
only introduce noise or redundancy and do not bring in useful information.
(3) \textbf{Our improved performance mainly comes from the diversified deformed
images, rather than the embedding sub-network.} Without producing
the deformed images in the meta-testing phase (`IDeMe-Net - Aug. Testing'),
the performance is close to the baseline, suggesting that training
on the deformed images does not obviously benefit from the embedding
sub-network. (4)
\textbf{Our meta-learned deformation sub-network effectively exploits
the complementarity and interaction between the probe and gallery
image patches, producing the key information in the deformed images.}
To show this point, we investigate two deformation strategies: randomly
generating the weight vector $w$ (`IDeMe-Net - Def. Network') and setting
all the weights to be 0 (`IDeMe-Net - Deform'); in the latter case, it
is equivalent to purely using the gallery images to serve as the deformed
images. Both strategies perform worse than the prototype classifier
baseline, indicating the importance of meta-learning a deformation
strategy.
 
\noindent \textbf{Different division schemes. }In the deformation
sub-network and Eq.~(\ref{eq:synthesize-1}), we evenly split the
image into $3\times3$ patches. Some alternative division schemes
are compared in Table~\ref{tab:resnet-10-imagenet1kablation} and Figure~\ref{fig:Ablation}(c).
Specifically, we consider the $1\times1$, $5\times5$, $7\times7$,
and pixel-level division schemes
and report the results as IDeMe-Net ($1\times1$), IDeMe-Net ($5\times5$),
IDeMe-Net ($7\times7$), and IDeMe-Net (pixel level), respectively.
The experimental results suggest the patch-level fusion, rather than
image-level or pixel-level fusion in our IDeMe-Net. The image-level
division ($1\times1$) ignores the local image structures  and deforms through
a global combination, thus decreasing the diversity. The pixel-level
division is particularly subject to the disarray of the local information,
while the patch-level division ($3\times3$, $5\times5$, and $7\times7$)
considers image patches as the basic unit to maintain some local information.
In addition, the results show that using a fine-grained patch size
(\eg, $5\times5$ division and $7\times7$ division) may achieve
slightly better results than our $3\times3$ division. In
brief, our patch-level division not only maintains the critical
region information but also increases diversity.

\noindent 
\begin{figure}
\centering 
\begin{centering}
\begin{tabular}{ccc}
\hspace{-0.3cm}\includegraphics[scale=0.2]{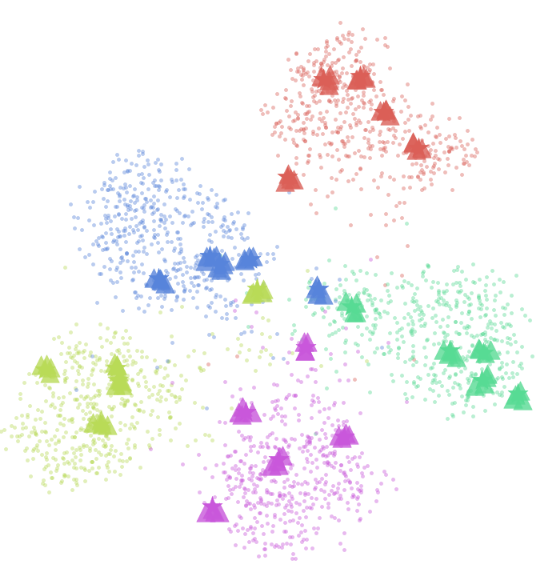}
\hspace{-0.4cm}  & \includegraphics[scale=0.2]{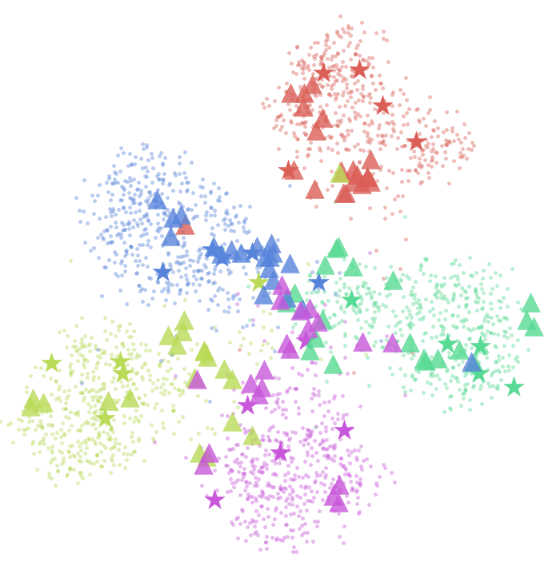} \hspace{-0.4cm}  & \includegraphics[scale=0.2]{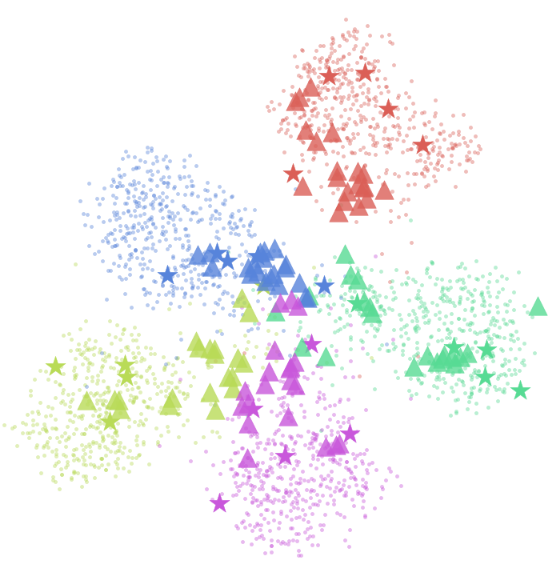}\tabularnewline
{\scriptsize{}{}(a) Gaussian Baseline }  & {\scriptsize{}{}(b) IDeMe-Net - Deform }  & {\scriptsize{}{}(c) IDeMe-Net}\tabularnewline
\end{tabular}
\par\end{centering}
\vspace{0.1in}
\caption{\label{fig:tsne}t-SNE visualization of 5 novel classes. Dots, stars, and triangles represent
the real examples, the probe images, and the synthesized deformed
images, respectively. (a) Synthesis by adding Gaussian noise. (b)
Synthesis by directly using the gallery images. (c) Synthesis by our
IDeMe-Net. \textbf{Best viewed in color with zoom.}}

 
\end{figure}

\noindent 
%
%
%

\begin{figure}[h]
	\begin{centering}
		\includegraphics[scale=0.45]{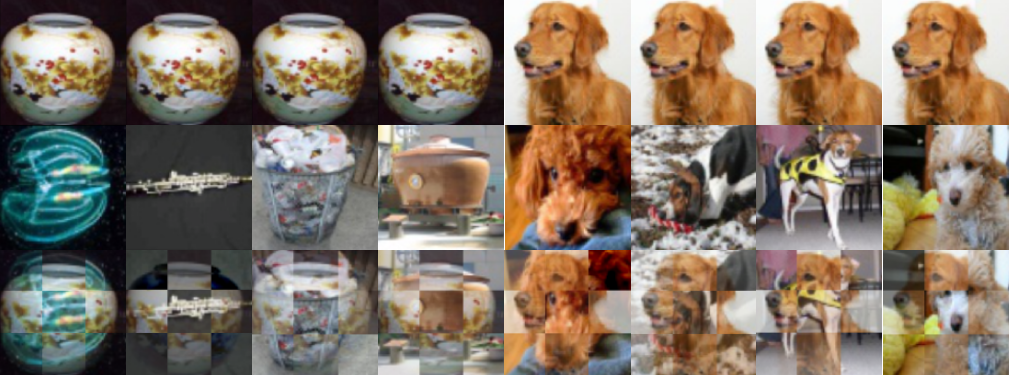}
		\par\end{centering}
\vspace{0.1in}
	\caption{\label{fig:example}{Examples of the deformed images during meta-testing. 1st row: probe images of {\em novel} classes. 2nd: gallery images of {\em base} classes. 3rd: synthesized images. The  probe-gallery image pairs from left to right: 
				vase--jellyfish, vase--oboe, vase--garbage bin, vase--soup pot, golden retriever--poodle, golden retriever--walker hound, golden retriever--walker hound, and golden retriever--poodle. \textbf{Best viewed in color with zoom.}
	}}
\end{figure}

\noindent \textbf{Number of synthesized deformed images.} We also
show how the top-5 accuracy changes with respect to the number of
synthesized deformed images in Figure~\ref{fig:Ablation}(d). Specifically,
we change the number of synthesized deformed images $n_{aug}$ in
the deformation sub-network, and plot the 5-shot top-5 accuracy
on the Imagenet 1K Challenge dataset. It shows that when $n_{aug}$ is changed
from $0$ to 8, the performance of our IDeMe-Net is gradually improved.
The performance saturates when enough deformed images
are generated ($n_{aug}>8$).

\noindent \textbf{Visualization of deformed images in feature space.} Figure~\ref{fig:tsne} shows the t-SNE
\cite{tsne} visualization of 5 novel classes from our IDeMe-Net,
the Gaussian noise baseline, and the `IDeMe-Net - Deform' variant. For the Gaussian noise
baseline, the synthesized images are heavily clustered and
close to the probe images. By contrast, the synthesized deformed images
of our IDeMe-Net scatter widely in the class manifold and tend to
locate more around the class boundaries. For `IDeMe-Net - Deform',
the synthesized images are the same as the gallery images
and occasionally fall into manifolds of other classes. Interesting,
comparing Figure~\ref{fig:tsne}(b) and Figure~\ref{fig:tsne}(c), our
IDeMe-Net effectively deforms those misleading gallery images back
to the correct class manifold.

\noindent \textbf{Visualization of deformed images in image space.} Here we show some examples of
our deformed images on novel classes in Figure~\ref{fig:example}. We can observe that the deformed images (in the third row) are visually different from the probe images (in the first row) and the gallery images (in the second row). For novel classes (\eg, vase and golden retriever), our method learns to find visual samples that are similar in shape and geometry (\eg, jelly fish, garbage bin, and soup pot) or similar in appearance (\eg, poodle and walker hound). By doing so, the deformed images preserve important visual content from the probe images and introduce new visual contents from the gallery images, thus diversifying and augmenting the training images in a way that maximizes the one-shot classification accuracy.



\subsection{Results on \emph{mini}ImageNet}


\begin{table}
\begin{centering}
{\small{}}%
\begin{tabular}{c|c|c}
\hline 
\multirow{2}{*}{{\small{}Method }} & \multicolumn{2}{l}{\emph{\small{}{}{}{}{}{}mini}{\small{}{}{}{}{}{}ImageNet
($\%$)}}\tabularnewline
\cline{2-3} 
 & {\small{}{}{}{}{}{}1-shot } & {\small{}{}{}{}{}{}5-shot }\tabularnewline
\hline 
\hline 
{\small{}{}{}{}{}{}MAML~\cite{MAML} } & {\small{}{}{}{}{}{}48.70\textpm 1.84 } & {\small{}{}{}{}{}{}63.11\textpm 0.92 }\tabularnewline
\hline 
{\small{}{}{}{}{}{}Meta-SGD~\cite{meta-sgd} } & {\small{}{}{}{}{}{}50.47\textpm 1.87 } & {\small{}{}{}{}{}{}64.03\textpm 0.94 }\tabularnewline
\hline 
{\small{}{}{}{}{}{}Matching Network~\cite{matchingnet_1shot} } & {\small{}{}{}{}{}{}43.56\textpm 0.84 } & {\small{}{}{}{}{}{}55.31\textpm 0.73 }\tabularnewline
\hline 
{\small{}{}{}{}{}{}Prototypical Network~\cite{prototype_network} } & {\small{}{}{}{}{}{}49.42\textpm 0.78 } & {\small{}{}{}{}{}{}68.20\textpm 0.66 }\tabularnewline
\hline 
{\small{}{}{}{}{}{}Relation Network~\cite{relation_net} } & {\small{}{}{}{}{}{}57.02\textpm 0.92 } & {\small{}{}{}{}{}{}71.07\textpm 0.69 }\tabularnewline
\hline 
{\small{}{}{}SNAIL~\cite{SNAIL} } & {\small{}{}{}{}55.71\textpm 0.99 } & {\small{}{}{}{}{}68.88\textpm 0.92}\tabularnewline
\hline 
{\small{}{}{}Delta-Encoder~\cite{Delta-encoder} } & {\small{}{}58.7 } & {\small{}{}{}73.6}\tabularnewline
\hline 
{\small{}{}{}Cos \& Att. \cite{dym}} & {\small{}{}{}{}{}{}55.45\textpm 0.89 } & {\small{}{}{}{}{}{}70.13 \textpm 0.68}\tabularnewline
\hline 
\hline 
{\small{}{}{}{}{}{}Prototype Classifier } & {\small{}{}{}{}{}{}52.54\textpm 0.81 } & {\small{}{}{}{}{}{}72.71\textpm 0.73}\tabularnewline
\hline 
{\small{}{}{}{}{}{}IDeMe-Net (Ours)} & {\small{}{}{}{}{}{}}\textbf{\small{}{}{}{}{}}{\small{}{}{}}\textbf{\emph{\noun{\small{}{}59.14}}}{\small{}{}{}{}{}\textpm 0.86 } & \textbf{\small{}{}{}{}{}{}74.63}{\small{}{}\textpm 0.74 }\tabularnewline
\hline 
\end{tabular}
\vspace{0.1in}
\caption{\label{tab:miniimagenet}\textbf{Top-1 accuracy (\%) on novel classes of the }\textbf{\emph{mini}}\textbf{ImageNet} \textbf{dataset.}
``\textpm '' indicates $95\%$ confidence intervals over tasks. }
\par\end{centering}
 
\end{table}

\noindent \textbf{Setup and Competitors.} We use a ResNet-18 architecture
as the embedding sub-network. We randomly sample 30 images per base
category to construct the gallery $G$. Other settings are the same as
those on the ImageNet 1k Challenge dataset. As summarized in Table~\ref{tab:miniimagenet},
we mainly focus on three groups of  competitors: (1) meta-learning
algorithms, such as MAML \cite{MAML} and Meta-SGD \cite{meta-sgd};
(2) metric learning algorithms, including matching networks \cite{matchingnet_1shot},
prototypical networks~\cite{prototype_network}, relation networks~\cite{relation_net}, SNAIL~\cite{SNAIL}, delta-encoder~\cite{Delta-encoder},
and Cosine Classifier \& Att. Weight Gen (Cos \& Att.)~\cite{dym}.

\noindent \textbf{Results. }We report the results in Table~\ref{tab:miniimagenet}. Impressively, our IDeMe-Net consistently outperforms all these
state-of-the-art competitors. This further validates the general effectiveness
of our proposed approach in addressing one-shot learning tasks.

\section{Conclusion}

In this paper, we propose a conceptually simple yet powerful approach
to address one-shot learning that uses a trained image deformation
network to generate additional examples. Our deformation network leverages
unsupervised gallery images to synthesize deformed images, which is
trained end-to-end by meta-learning. The extensive experiments demonstrate
that our approach achieves state-of-the-art performance on multiple
one-shot learning benchmarks, surpassing the competing methods
by large margins.
\\
\\
\indent \textbf{Acknowledgment:} This work is supported in part by the grants from NSFC (\#61702108),   STCSM (\#16JC1420400), Eastern Scholar (TP2017006), and The Thousand Talents Plan of China (for young professionals, D1410009).

 \bibliographystyle{ieee}
\bibliography{ref_updated}


\end{document}